\documentclass[letterpaper, 10 pt, conference]{ieeeconf}  %

\IEEEoverridecommandlockouts                              %

\usepackage{graphicx}
\usepackage{subcaption}
\usepackage[export]{adjustbox}

\usepackage{amssymb}
\usepackage{amsmath}
\usepackage{commath}
\usepackage{mathtools}

\usepackage[pdfborder={0 0 0}]{hyperref}
\usepackage{import}
\usepackage{paralist}
\usepackage{todonotes}

\DeclareMathOperator{\sech}{sech}

\title{\LARGE \bf
UKF-Based Sensor Fusion \\ for Joint-Torque Sensorless Humanoid Robots}

\author{Ines Sorrentino$^{1,2}$, Giulio Romualdi$^{1}$, Daniele Pucci$^{1,2}$%
\thanks{*This work was not supported by any organization}%
\thanks{${}^{1}$ Artificial and Mechanical Intelligence, Italian Institute of Technology, Genoa, Italy, {\tt\small (e-mail: name.surname@iit.it)}}
\thanks{${}^{2}$ Machine Learning and Optimisation, University of Manchester, Manchester, UK}%
}

\begin{document}

\maketitle
\thispagestyle{empty}
\pagestyle{empty}

\begin{abstract}

This paper proposes a novel sensor fusion based on Unscented Kalman Filtering for the online estimation of joint-torques of humanoid robots without joint-torque sensors.  At the feature level, the proposed approach considers multi-modal measurements (e.g. currents, accelerations, etc.) and non-directly measurable effects, such as external contacts, thus leading to joint torques readily usable in control architectures for human-robot interaction. The proposed sensor fusion can also integrate distributed, non-collocated force/torque sensors, thus being a flexible framework with respect to the underlying robot sensor suit. To validate the approach, we show how the proposed sensor fusion can be integrated into a two-level torque control architecture aiming at task-space torque-control. 
The performances of the proposed approach are shown through extensive tests on the new humanoid robot ergoCub, currently being developed at Istituto Italiano di Tecnologia. We also compare our strategy with the existing state-of-the-art approach based on the recursive Newton-Euler algorithm. Results demonstrate that our method achieves low root mean square errors in torque tracking, ranging from 0.05 Nm to 2.5 Nm, even in the presence of external contacts.

\end{abstract}

\section{Introduction}
\label{sec:introdcution}

Humanoid robots have emerged as versatile tools with applications ranging from assistive devices to complex human-robot interactions. Precise control of these robots is crucial for replicating human-like movements and ensuring safe interactions in dynamic environments~\cite{9035060, romualdi2020benchmarking, dafarra2022dynamic}. 
Among the control strategies, torque control has gained significant attention for effectively operating in unpredictable situations, providing robust motion tracking, and ensuring compliance with unexpected contacts~\cite{mesesan2019dynamic,ramuzat2021comparison, de2005sensorless}. Nevertheless, the implementation of this control strategy requires knowledge of the effective joint torques. When joint torque sensors are unavailable, the accurate estimation of these torques becomes a fundamental requirement to achieve high performance in torque tracking and dynamic tasks. This paper proposes a joint torque estimation strategy to enable torque control on sensorless humanoid robots to guarantee compliance to unexpected disturbances.

\begin{figure}[ht]
    \raggedleft
    \begin{subfigure}[b]{0.35\linewidth}
        \raggedright
        \includegraphics[width=\linewidth]{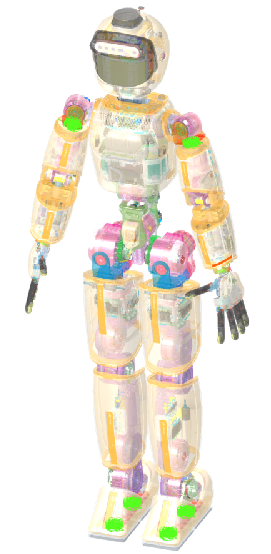}
        \label{fig:trasp}
    \end{subfigure}
    \hfill
    \begin{subfigure}[b]{0.57\linewidth}
        \raggedleft
        \includegraphics[width=\linewidth]{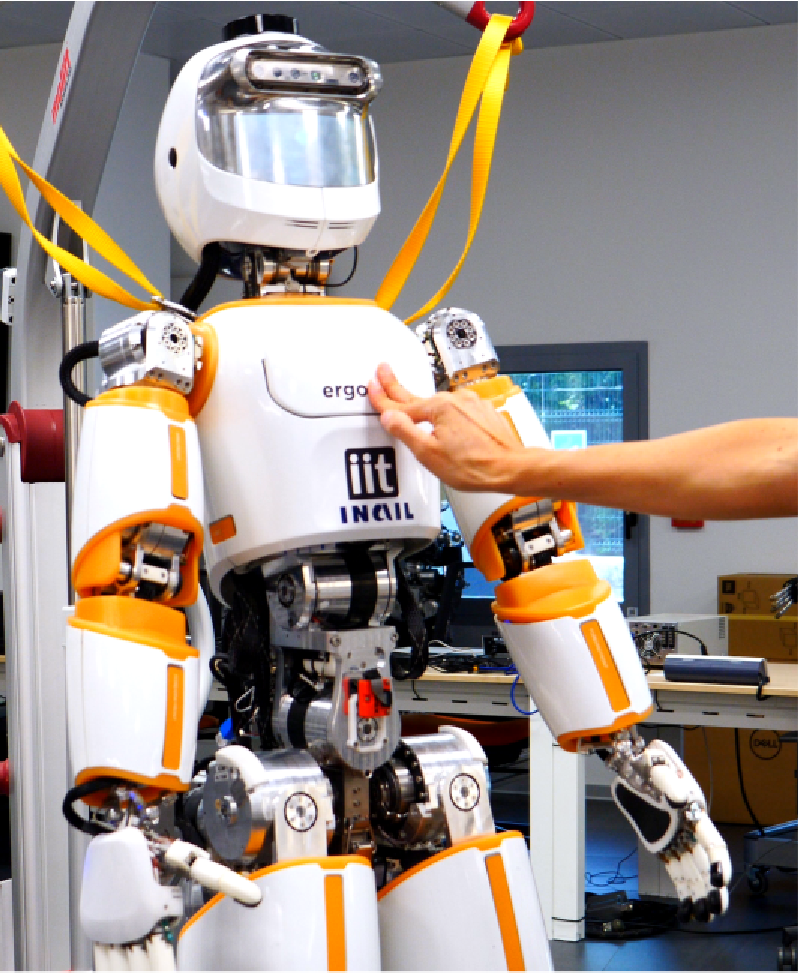}
        \label{fig:pushed}
    \end{subfigure}
    \caption{On the left The ergoCub humanoid robot in transparency with FT sensors highlighted in green. On the right ergoCub being pushed on the torso during a closed-loop torque controller test.}
    \label{fig:ergocub}
    \vspace{-10pt}
\end{figure}

Many implementations of torque-based whole-body control algorithms have been developed for robots equipped with joint torque sensors~\cite{ramuzat2021comparison,henze2016passivity,ferrari2023multi}.
However, integrating these sensors into humanoid robots presents challenges due to space constraints, costs, mechanical complexity, and potential problems like measurement drift and joint elasticity~\cite{romualdi2022whole,min2019novel}.
Consequently, numerous research teams have designed their robot platforms without joint torque sensors. In these scenarios, torque control is often replaced by a stiff position control strategy, and dynamic tasks are satisfied under specific assumptions based on the application, such as flat terrain or absence of obstacles~\cite{romualdi2018benchmarking,romualdi2022online,hondawalking}. To overcome these limitations and enable joint torque control on such robots, the challenge is accurately estimating the joint torques.

One of the most common approaches for torque estimation on humanoid robots is based on the classical recursive Newton-Euler algorithm (RNEA). The RNEA approach uses the robot dynamic model to propagate the measured wrenches from the force/torque (FT) sensors along the robot kinematic chain. However, this deterministic estimation method neglects effects like measurement noises, miss-calibrated sensors, or modeling uncertainties. Consequently, torque estimation often does not achieve the desired performance levels, especially when dealing with external contacts, which can only influence the estimate when applied after the FT sensors in the kinematic chain~\cite{del2016implementing,fumagalli2012force,traversaro2015inertial}. %

A more recent approach overcomes the problem of considering external contacts by estimating joint torques from the motor current, motor angular velocity, and motor angular acceleration measured in each joint~\cite{nagamatsu2017distributed}. This method remains deterministic and relies on prior knowledge of certain coefficients, including the motor torque constant, motor moment of inertia, and the coefficient of friction for the harmonic drives. These coefficients have a direct impact on the algorithm outcomes, but measuring them can be challenging. Additionally, the deterministic nature of this approach overlooks modeling uncertainties, and the accuracy of joint torque estimation depends entirely on how effectively the model identification procedure is executed.

This paper contributes towards the design of a novel joint torque estimator with the aim of enabling torque control on humanoid robots without joint torque sensors and mounting high-ratio harmonic drives. More precisely, unlike classical deterministic algorithms, our proposed methodology is based on Unscented Kalman Filetering~\cite{wan2001unscented} and sensor fusion~\cite{hackett1990multi} to combine data acquired from distributed sensors, including joint encoders, FT sensors, inertial sensors, and motor current sensors. This approach is employed to address non-measurable factors, such as external contacts, modeling uncertainties, and measurement noise.
The joint torque estimation algorithm is tested in combination with a two-layers torque control architecture to track references in the joint or cartesian spaces. The high-level control law is implemented as a constrained quadratic programming problem and generates the reference joint torques. The low-level torque controller is implemented as a Proportional-Integral (PI) feedback-feedforward control with a friction compensation action to reduce energy losses. Simulations and experiments on the humanoid robot ergoCub (see Fig. \ref{fig:ergocub}) demonstrate the effectiveness of our approach. A comparison with the state-of-the-art RNEA algorithm for torque estimation shows improvements in terms of estimation accuracy, in particular in the presence of external contacts.

\begin{figure*}[t]
    \centering
    \includegraphics[width=340pt]{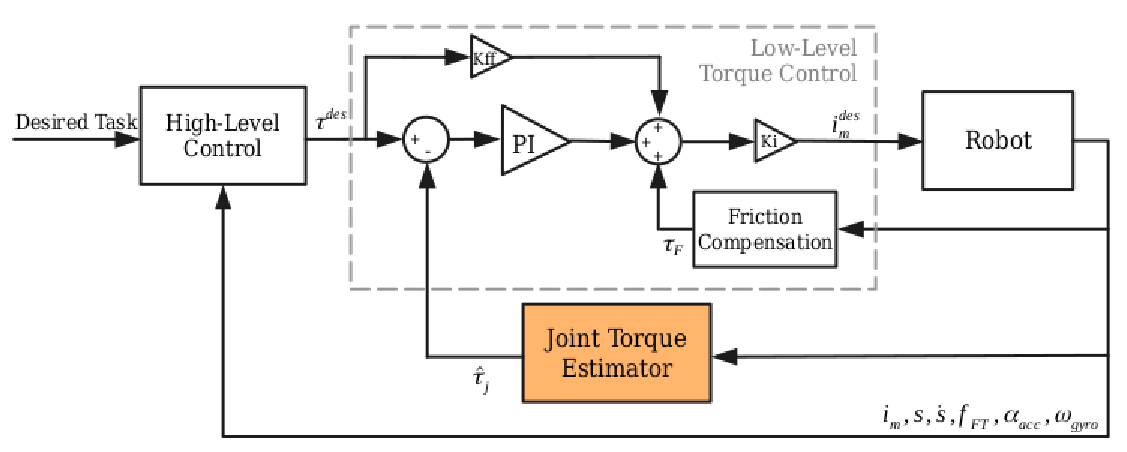}
    \caption{Block diagram of the controller architecture implemented on the ergoCub humanoid robot.}
    \vspace{-9pt}
    \label{fig:controlarchitecture}
\end{figure*}

The paper is organized as follows. Sec.~\ref{sec:background} introduces the notation and recalls some concepts of floating-base systems. Sec.~\ref{sec:RDE} details the joint torque estimation. Sec.~\ref{sec:experiment} presents the validation results on the position-controlled humanoid robot ergoCub. Sec.~\ref{sec:conclusions} concludes the paper.

\section{Background}
\label{sec:background}

\subsection{Notation}
\begin{itemize}
\item $I_{m \times n}$ and $0_{m \times n}$ denote the $m \times n$ identity and zero matrices respectively; when $m=n$ we use simply $I_{n}$ and $0_{n}$.
\item $\mathcal{I}$ denotes the inertial frame.
\item $\prescript{\mathcal{I}}{}{p}_\mathcal{B}$ is a vector connecting the origin of frame $\mathcal{I}$ and the origin of frame $\mathcal{B}$ expressed in frame $\mathcal{I}$.
\item given $\prescript{\mathcal{I}}{}{p}_\mathcal{B}$ and $\prescript{\mathcal{B}}{}{p}_\mathcal{C}$,  $\prescript{\mathcal{I}}{}{p}_\mathcal{C} = \prescript{\mathcal{I}}{}{R}_\mathcal{B} \prescript{\mathcal{B}}{}{p}_\mathcal{C} + \prescript{\mathcal{I}}{}{p}_\mathcal{B}= \prescript{\mathcal{I}}{}{H}_\mathcal{B} \begin{bmatrix}
  \prescript{\mathcal{B}}{}{p}_\mathcal{C} ^\top & \;1
\end{bmatrix}^\top$. $\prescript{\mathcal{I}}{}{H}_\mathcal{B}$ is the homogeneous transformation and $\prescript{\mathcal{I}}{}{R}_\mathcal{B} \in SO(3)$ is the rotation matrix.
\item $\scalebox{0.8}{$\prescript{\mathcal{B}}{}{\textrm{v}}_\mathcal{I,B} = \begin{bmatrix} \prescript{\mathcal{B}}{}{v}_\mathcal{I,B} ^\top & \prescript{\mathcal{B}}{}{\omega}_\mathcal{I,B}^\top \end{bmatrix}^\top \in \mathbb{R}^6$}$ is the velocity of the frame $\mathcal{B}$ with respect to the frame $\mathcal{I}$ expressed in the frame $\mathcal{B}$, where $\prescript{\mathcal{B}}{}{v}_\mathcal{I,B}$ denotes the linear velocity and $\prescript{\mathcal{B}}{}{\omega}_\mathcal{I,B}$ the angular velocity. For the sake of simplicity, in the rest of the paper, we use ${\textrm{v}}_\mathcal{B}$.
\item $\times$ is the cross product on $\mathbb{R}^3$.
\item $g_\mathcal{B}$ is the gravity vector expressed in $\mathcal{B}$.
\item $\scalebox{0.8}{$\alpha_{\mathcal{S}} = \prescript{\mathcal{S}}{}{\dot{\textrm{v}}_{I,S}} - \begin{bmatrix} (\prescript{\mathcal{S}}{}{R_\mathcal{I}}\prescript{I}{}{v_\mathcal{S}} ) \times \prescript{\mathcal{\mathcal{S}}}{}{\omega_{\mathcal{I},\mathcal{S}}} \\ 0_{3 \times 1} \end{bmatrix}$}$ is the conversion rule between the acceleration of a sensor with frame $\mathcal{S}$ %
with respect to the inertial frame and the sensor acceleration $\alpha_{\mathcal{S}}$ expressed in the sensor frame.
\item $i_m \in \mathbb{R}^n$ is the vector of motor currents with $n$ the number of motors. 
\item  $\scalebox{0.8}{$\prescript{}{}{\mathrm{f}}_\mathcal{B}^ \top = \begin{bmatrix} {{f}}_\mathcal{B} ^ \top & {\mu}_\mathcal{B}^ \top \end{bmatrix}$}$ is the wrench acting on a point of a rigid body expressed in the frame $\mathcal{B}$.
\item the \textit{hat symbol} identifies the variables estimated through the Unscented Kalman Filter.
\item For clarity, all the sensor measurements will be expressed in the sensor frames.
\end{itemize}

\subsection{Humanoid Robot Model}
A humanoid robot is a floating base multi-body system composed of $n+1$ links connected by $n$ joints with one degree of freedom each. The configuration of the robot is defined by considering both the joint positions $s$ and the homogeneous transformation from the inertial frame to the robot base frame $\mathcal{B}$. The configuration is identified by the triplet $q = (\prescript{\mathcal{I}}{}{p}_\mathcal{B}, \prescript{\mathcal{I}}{}{R}_\mathcal{B}, s) \in  \mathbb{R}^3 \times SO(3) \times \mathbb{R}^n$. The velocity of the system is characterized by the set $\nu = (\prescript{\mathcal{I}}{}{v}_\mathcal{B}, \prescript{\mathcal{I}}{}{\omega}_\mathcal{B}, \dot{s})$ where $\prescript{\mathcal{I}}{}{\omega}_\mathcal{B}$ is the angular velocity of the base frame expressed with respect to the inertial frame, $\dot{s}$ is the time derivative of the joint positions.

The equation of motion of a multibody system is described by applying the \textit{Euler-Poincar\'e Formalism}~\cite{featherstone2014rigid}
\begin{equation}
M(q) \dot \nu + h(q, \nu) = B \tau_j + \sum_{j} {J_j(q)^\top f_{ext,j}} \; ,
\label{eq:robotdynamics}
\end{equation}
where $M \in \mathbb{R}^{(6+n) \times (6+n)}$ is the mass matrix, $h(q, \nu) \in \mathbb{R}^{6+n}$ accounts for the Coriolis, centrifugal and gravitational effects, $\tau_j \in \mathbb{R}^{n}$ are the joint torques, $B \in \mathbb{R}^{(6+n) \times n}$ is the selection matrix, $f_{ext,j} \in \mathbb{R}^6$ is the $j$-th contact wrench expressed in the contact frame, $J_j(q)$ is the Jacobian associated to the $j$-th contact wrench.

\subsection{Model of the harmonic drive mechanical friction}
Humanoid robots are often equipped with harmonic drives having high reduction ratios~\cite{gandhi2002modeling, ismail2021simplified}. The performance of harmonic drives is influenced by significant non-linear effects such as kinematic error, flexibility, hysteresis, and friction~\cite{cisneros2020reliable,gandhi2002modeling}. Among these, friction is the predominant factor responsible for dissipating a large part of the torque generated by the input current to the motors. Consequently, it significantly influences the accuracy of joint torque estimation. The effective joint torque can be calculated as
\begin{equation}
\label{eq:frictionfromresidual}
\tau_j = r \tau_m - \tau_F \; ,
\end{equation}
where $r$ is the gear-ratio, $\tau_m$ is the motor torque, and $\tau_F$ is the friction torque. We approximate the friction model through the combination of Coulomb and viscous friction models. To avoid discontinuities at zero velocity, we introduce the hyperbolic tangent function into our model~\cite{johanastrom2008revisiting}:
\begin{equation}\label{eq:friction}
\tau_F = k_0  \tanh(k_1 \dot s) + k_2 \dot s \; .
\end{equation}
The identification of the coefficients $k_0$, $k_1$, $k_2$ is discussed in Section \ref{sec:experiment}.

\section{UKF Based Stochastic Sensor Fusion for Joint Torques}
\label{sec:RDE}

In this section, we describe the estimation algorithm for the robot joint torques $\tau_j$, by combining data from the physical sensors on the robot through sensor fusion. Because the process and measurement models are non-linear, we formulate this problem as an Unscented Kalman Filter~\cite{wan2001unscented}.
Unlike traditional approaches that require linearization~\cite{shi2014torque}, the UKF provides accurate state estimates without requiring this linearization step. Nonetheless, the UKF still depends on statistical information regarding the noise affecting both the state and measurements, specifically their covariance. To achieve the best performance, it is crucial to fine-tune these covariances.

\subsection{Unscented Kalman Filter definition}
We exploit a Kalman Filter to fuse the sensor data and estimate the system state. The discrete-time nonlinear dynamic system is written as
\begin{IEEEeqnarray}{cl}
\IEEEnonumber
x_{k+1} &= F(x_k,u_k,v_k) \\
y_k &= G(x_k,u_k,w_k)
\end{IEEEeqnarray}
\vspace{-10pt}
where %
\begin{IEEEeqnarray}{cl}
\IEEEnonumber
& x_k = \begin{bmatrix}\hat{\dot s}^\top & \hat{\tau}_{m}^\top & \hat{\tau}_{F}^\top & \hat{f}_{FT}^\top  & \hat{f}_{ext}^\top \end{bmatrix} ^\top_k \\
& y_k = \begin{bmatrix}\dot{s}^\top & i_{m}^\top & f_{FT}^\top & \alpha_{acc}^\top & \omega_{gyro}^\top \end{bmatrix}^\top_k  \\ 
\IEEEnonumber
&u_k =\begin{bmatrix}H_{\mathcal{B}}^\top & {\textrm{v}}_{\mathcal{B}}^\top & {\dot{\textrm{v}}}_{\mathcal{B}}^\top & s^\top \end{bmatrix}^\top_k \; .
\end{IEEEeqnarray}
where $\hat{\dot s}$, $\hat{\tau}_{m}$, $\hat{\tau}_{F}$, $\dot{s}$, $i_{m}$ $\in \mathbb{R}^n$, with $n$ the number of joints. $\hat{f}_{FT}$, $f_{FT}$ $\in \mathbb{R}^{6m}$ where $m$ is the number of FT sensors. $\hat{f}_{ext} \in \mathbb{R}^{6l}$ where $l$ is the number of external contacts. $\alpha_{acc} \in \mathbb{R}^{3p}$ is the vector of the linear accelerations measured by the accelerometers, being $p$ the number of accelerometers. $\omega_{gyro} \in \mathbb{R}^{3h}$ is the vector of all the angular velocities measured by the gyroscopes, being $h$ the number of gyroscopes.
The process noise $v_k$ drives the dynamical system, while $w_k$ is the observation noise.
It is worth noting that, $F$ and $G$ are non-linear equations. 
 
\subsection{Process Model}
The list of dynamic equations composing the process model of the UKF is defined as follows.
\begin{itemize}
    \item[-] \textit{Joint velocity dynamics}: we split the robot model at the FT sensors shown in green in Fig. \ref{fig:ergocub}. Then, considering Eqs. \eqref{eq:robotdynamics} and \eqref{eq:frictionfromresidual} we define the dynamic equation of each submodel $i$
    \begin{IEEEeqnarray}{cl}
    \IEEEnonumber
    \label{eq:FD}
    \dot \nu^{(i)} = {M(q)^{-1} }^{(i)}  \big(
        & r \hat{\tau}_m^{(i)} {-} \hat{\tau}_F^{(i)} {-} h(q, \nu)^{(i)} \\ &{+} \sum_{j}J_{FT,j}^\top \hat{f}_{FT,j} {+} J_{ext}^\top \hat{f}_{ext}
    \big) \; ,
    \end{IEEEeqnarray}
    This allows us to treat FT measurements as external contacts for the sub-models. From $\dot \nu^{(i)}$ we select the joint velocities $\dot{s}^{(i)}$ of the submodel $i$.
    \item[-] \textit{Friction torque dynamics}: the dynamics of the friction torque is computed as the time derivative of equation~\eqref{eq:friction} 
    \begin{equation}\dot{\hat{\tau}}_F = \big( k_0 k_1 \sech^2 ( k_1 \dot s ) + k_2 \big) \ddot s
    \end{equation}
    where $\ddot s$ is computed from the forward dynamics \eqref{eq:FD}. 
    \item[-] \textit{Motor torque, FT sensor, and contact wrench dynamics}: the dynamics of the motor torques, FT sensors, and contact wrenches are unknown, and we consider a constant dynamic model $\dot{\hat{\tau}}_m  = 0_{n \times 1}$, $\dot{\hat{f}}_{FT} = 0_{6m \times 1}$, and  $\dot{\hat{f}}_{ext} = 0_{6l \times 1}$.
\end{itemize}

\subsection{Measurement Model}
The measurement model accounts for all the sensors mounted on the robot, implementing a sensor fusion approach. %
\begin{itemize}
    \item[-] \textit{Joint velocity measurement}: $\dot s = \dot{\hat{s}}$.
    \item [-] \textit{Motor current measurement}: the current sensor measurement is related to the state vector as $ i_m =  k_{\tau}^{-1}  \hat{\tau}_m $ where $k_{\tau}$ is the motor torque constant.
    \item[-] \textit{FT sensor measurement}: $f_{FT}= \hat{f}_{FT}$.
    \item[-] \textit{Accelerometer sensor measurement}: the accelerometer measures the linear accelerations expressed in the sensor frame. The measurement model of an accelerometer is defined as:
    \begin{IEEEeqnarray}{cl}
    \IEEEnonumber
        \alpha_{acc} = \big( \dot{J} \nu + J \dot{\nu} & \big)_{3 \times 1} - \prescript{\mathcal{S}}{}{R}_\mathcal{A} \prescript{\mathcal{A}}{}{g} \\ & - (\prescript{\mathcal{S}}{}{R_\mathcal{A}}\prescript{A}{}{\dot o_\mathcal{S}} ) \times \prescript{\mathcal{\mathcal{S}}}{}{\omega_{\mathcal{A},\mathcal{S}}} \; .
    \end{IEEEeqnarray}
    where all the variables on the right depend on the state $\dot{\hat{s}}$ and the input $s$. The $3 \times 1$ subscript means that we select the first three elements of the resulting vector.
    \item[-] \textit{Gyroscope sensor measurement}: the gyroscope measures the angular velocities expressed in the sensor frame, and the measurement model is defined by selecting the angular part $\omega_{gyro}$ of the equation 
      $  \textrm{v}_{gyro} = J \nu $ where $J$ and $\nu$ depend on the state $\dot{\hat{s}}$ and the input $s$.
\end{itemize}

\subsection{Covariance tuning}
We have performed the following trial-and-error procedure to fine-tune the covariances of the UKF observer. We have generated joint torque trajectories using the high-level controller described in Section \ref{sec:HLC} which have been used as references for the ergoCub robot simulated in Gazebo~\cite{gazebosimGazebo}. The joint positions measured during the simulation are adopted as reference trajectories for the joints of the real robot. While executing the trajectory, we log data from all sensors, creating offline datasets to be used as input and measurements for the UKF.
This process allows us to tune the process and measurement covariances, ensuring that the UKF accurately estimates joint torques, matching the joint torques coming from the simulation.

\section{Experiments}
\label{sec:experiment}

This section presents the validation results of the joint torque estimation method introduced in Sec. \ref{sec:RDE}. We also describe the controller architecture used for validation and the friction identification procedure. The code is available online~\cite{code}. We test our algorithm on the ergoCub humanoid robot (Fig. \ref{fig:ergocub}) equipped with various sensors including $19$ joint encoders, $19$ motor current sensors, $6$ FT sensors placed $2$ under each foot and $1$ on each arm respectively, and $7$ accelerometers and gyroscopes placed $2$ under each foot, $1$ on each arm, $1$ on the pelvis. The joint torque estimator is seamlessly integrated into the robot’s computer, activating upon startup. The results reported in the paper are based on data collected at various time points post-activation, demonstrating robust performance with low tracking errors, even after prolonged usage.

Validation experiments involve placing the robot on a pole to simplify the high-level control architecture, as the necessity to maintain balance is irrelevant to the study. Three experiments are conducted. In two of them, the proposed strategy is compared with a state-of-the-art joint torque estimation based on RNEA~\cite{traversaro2015inertial} to analyze reactions to external disturbances. In the last experiment, we analyze the performance of our method with a more complex task in the cartesian space. All the tests are shown in the supplementary video.

\subsection{Controller architecture}
\label{sec:controller}

\begin{figure*}[t]
\centering
\begin{subfigure}{.5\textwidth}
  \includegraphics[width=0.95\linewidth,right]{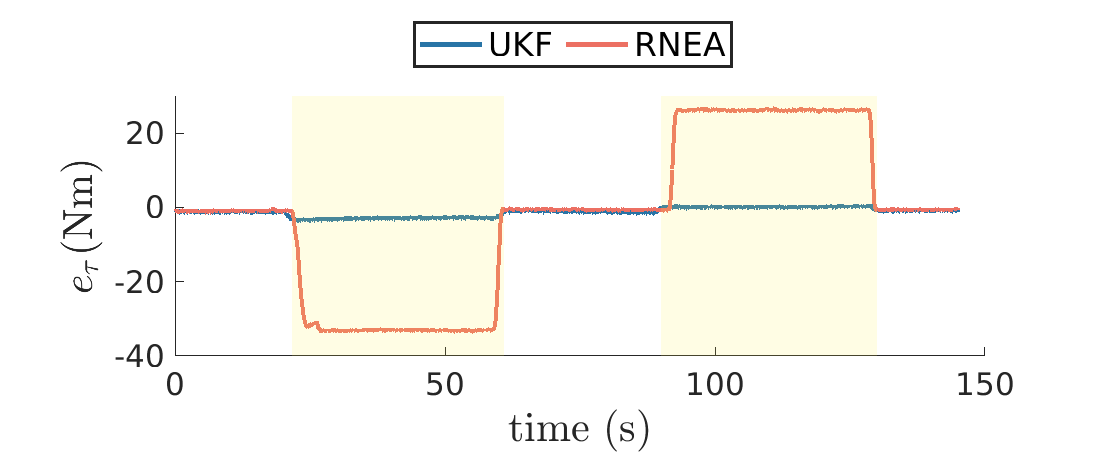}
  \caption{Zero torque reference tracking.}
  \label{fig:zerotrq}
\end{subfigure}%
\begin{subfigure}{.5\textwidth}
  \includegraphics[width=0.95\linewidth,left]{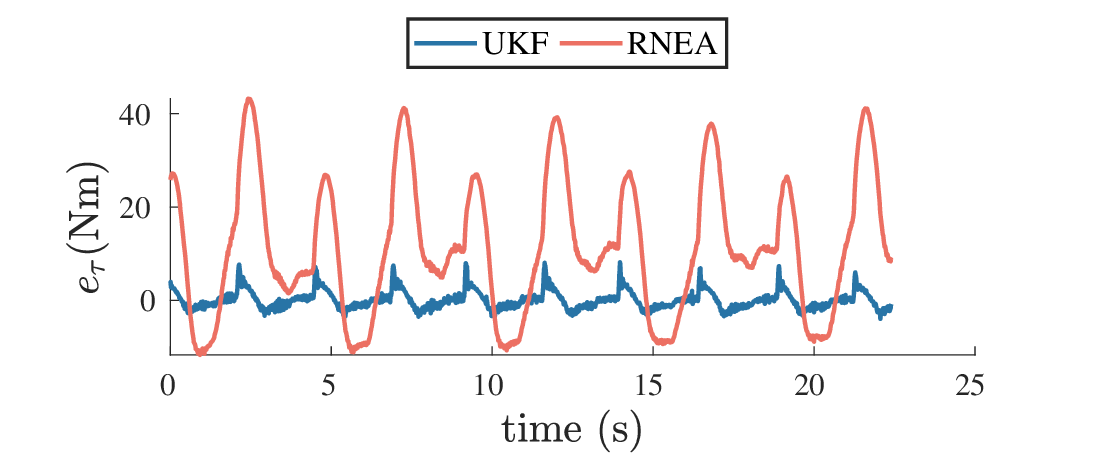}
  \caption{PD control with gravity compensation.}
  \label{fig:pdobstacle}
\end{subfigure}
\caption{(a)-(b) Comparison of the joint torque tracking errors obtained with the RNEA and UKF algorithms.}
\end{figure*}

\begin{figure*}[t]
    \centering
    \includegraphics[width=0.95\linewidth]{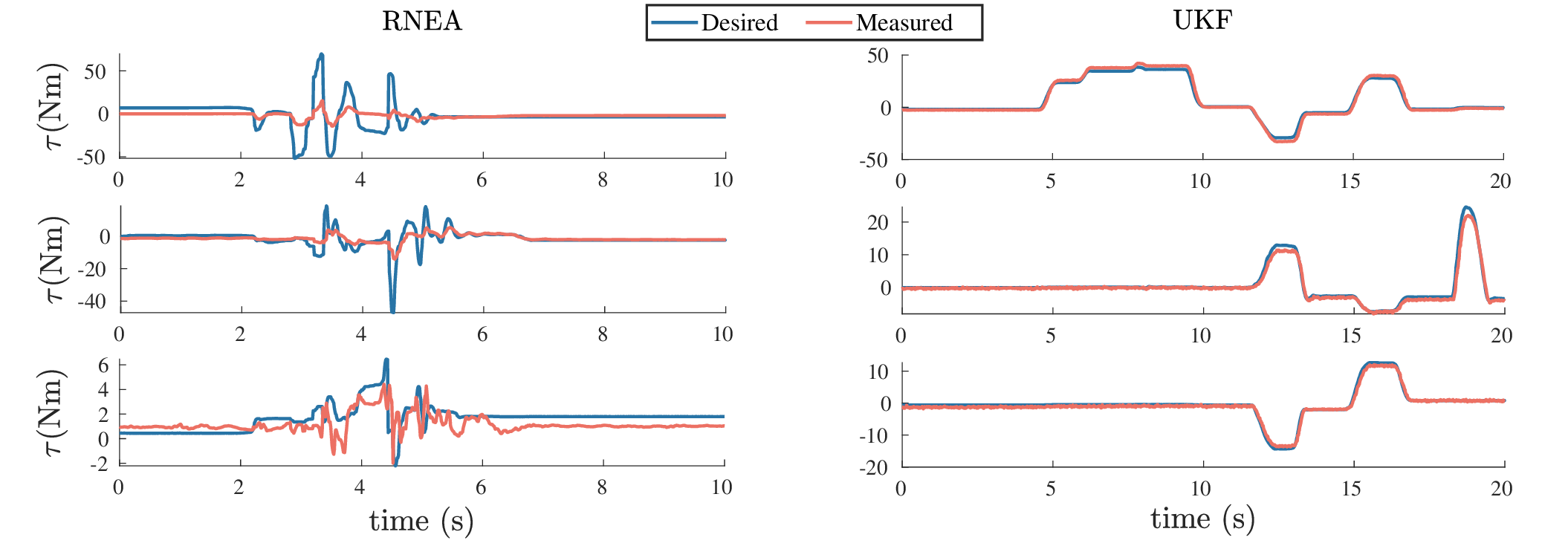}
    \caption{Comparison of the joint torque tracking for the PD control with gravity compensation experiment on torso joints. On the left are the results obtained with the RNEA algorithm, and on the right results given by the UKF method.}
    \vspace{-8pt}
    \label{fig:torso}
\end{figure*}

The control architecture is shown in Figure \ref{fig:controlarchitecture} and is detailed below.

\subsubsection{High-Level Control}
\label{sec:HLC}
Based on the robot dynamics \eqref{eq:robotdynamics}, the high-level control law computes the desired joint torques $\tau_{des}$ solving a constrained quadratic programming problem~\cite{romualdi2022whole}. The tasks are handled as cost functions and constraints. This section introduces the list of tasks considered in the control problem.

\begin{itemize}
\item[-] Cartesian task:
We account for desired link position and orientation with respect to the inertial frame $\prescript{\mathcal{A}}{}{H}^{des}_L = ( p^{des}_{L}, \prescript{\mathcal{A}}{}{R}^{des}_L ) \in \mathbb{R}^3 \times SO(3) $. The task is defined at the acceleration level as 
$\scalebox{0.8}{$ \Psi_{L_{SE(3)}} = \begin{bmatrix} \dot{v}^{*^\top}_{L} & {}^\mathcal{A} \dot{\omega}_{L}^{*^\top} \end{bmatrix}^\top - J_L \dot{\nu} - \dot{J}_L \nu $}$
where $\dot{v}^{*}_{L}$ and $\prescript{\mathcal{A}}{}{\dot{\omega}^{*}_L}$ are set to guarantee almost global stability.

\item[-] Joint position regularization task: This task prevents huge variations of the desired joint accelerations. It is described by \mbox{$ \Psi_{s} = \ddot{s}^{*} - \begin{bmatrix} 0_{n \times 6}  & I_{n}  \end{bmatrix} \dot{\nu}$}.
$\ddot{s}^{*}$ is defined as $\scalebox{0.8}{$
    \ddot{s}^{*} = \ddot{s}^{des} + k^d_s (\dot{s}^{des} - \dot{s}) + k^p_s (s^{des} -s)
$}$
where $s^{des}$ is the desired joint trajectory, and $k^d_s$ and $k^p_s$ are two positive-defined diagonal matrices.
\end{itemize}

\subsubsection{Low-Level Torque Control}

The joint torques $\tau^{des}$ generated by the high-level controller described in Section \ref{sec:HLC}, are sent to a decentralized low-level joint torque controller. The torque controller, shown in Figure \ref{fig:controlarchitecture}, is implemented as a feedback-feedforward control and includes the compensation of the mechanical friction introduced in Section \ref{sec:background} to mitigate energy losses~\cite{siciliano2010robotics}. The control law generates the reference currents to drive the motors.

\subsection{Friction parameters identification}
The identification of the friction models showed the presence of substantial static and viscous friction between the motor and the load. We have generated reference motor currents to drive the motors and identify the friction model associated with each harmonic drive. Under conditions of no external disturbances, the residual between the torque applied to the motors and the torque at the joint, as calculated in \eqref{eq:frictionfromresidual}, corresponds to the friction forces outlined in equation \eqref{eq:friction}. The friction parameters for the leg joints are reported in Table \ref{tab:friction}.
\begin{table}[]
\centering
\begin{tabular}{|l|c|c|c|}
\hline
& $k_0$ (Nm) & $k_1$ (1/rad/sec) & \multicolumn{1}{l|}{$k_2$ (Nm/rad/sec)} \\ \hline
hip pitch   & 4.9            & 4.0                   & 0.6                                         \\ \hline
hip roll    & 4.0            & 4.7                   & 0.3                                         \\ \hline
hip yaw     & 2.5            & 2.6                   & 0.5                                         \\ \hline
knee        & 2.3            & 2.7                   & 0.1                                         \\ \hline
ankle pitch & 2.3            & 2.3                   & 0.3                                         \\ \hline
ankle roll  & 1.3            & 2.0                   & 0.3                                         \\ \hline
\end{tabular}
\caption{Friction parameters for ergoCub leg joints.}
\label{tab:friction}
\end{table}
From experimental analysis, we assume negligible friction affecting the torso joints. 

\subsection{Comparison with the state-of-the-art}
We conducted a comparative analysis between our approach and the state-of-the-art method, which relies on RNEA. The setup of our ergoCub robot differs from that of the iCub robot used in the studies proposed in the state-of-the-art~\cite{del2016implementing} only in terms of the location of the FT sensors. Indeed, the FT sensors in the upper part of the legs are absent in the ergoCub robot. To ensure a fair comparison of the performance of the two algorithms under the same working conditions, we focused also on the torso chain which is cut at the FT sensor of the upper arms and the FT sensors of the feet. However, given that the robot is mounted on a pole, the legs do not influence the estimation. The comparison is conducted through two separate experiments, detailed in the following subsections. Both the RNEA and UKF approaches utilize identical high-level and low-level controller gains.

\subsubsection{Zero Torque Tracking}
In this experiment, we focus on controlling one joint in the robot leg to track a zero-torque reference, while other leg joints are position-controlled. An external force, generated by contact with an operator, perturbs the leg links and causes a change in the joint configuration. The external force is not measurable by the FT sensors being located lower in the kinematic chain. Consequently, the classical RNEA is unable to account for this contact in the estimation process. In Figure \ref{fig:zerotrq}, we compare the RNEA algorithm (in red) and our UKF method (in blue) in tracking the desired torque ($\tau^{des}=0$ Nm). The yellow areas represent the contact phases. RNEA estimates the joint torque only based on leg configuration, resulting in a significant error and causing the controller to react oppositely to the external force direction. Our approach uses motor current measurements to accurately estimate the joint torque with small error (approximately $2.5$ Nm), improving control performance and allowing for smoother movement through different joint configurations. Similar enhancements are observed in other leg joints.

\begin{figure*}[t]
    \centering
    \includegraphics[width=0.95\linewidth]{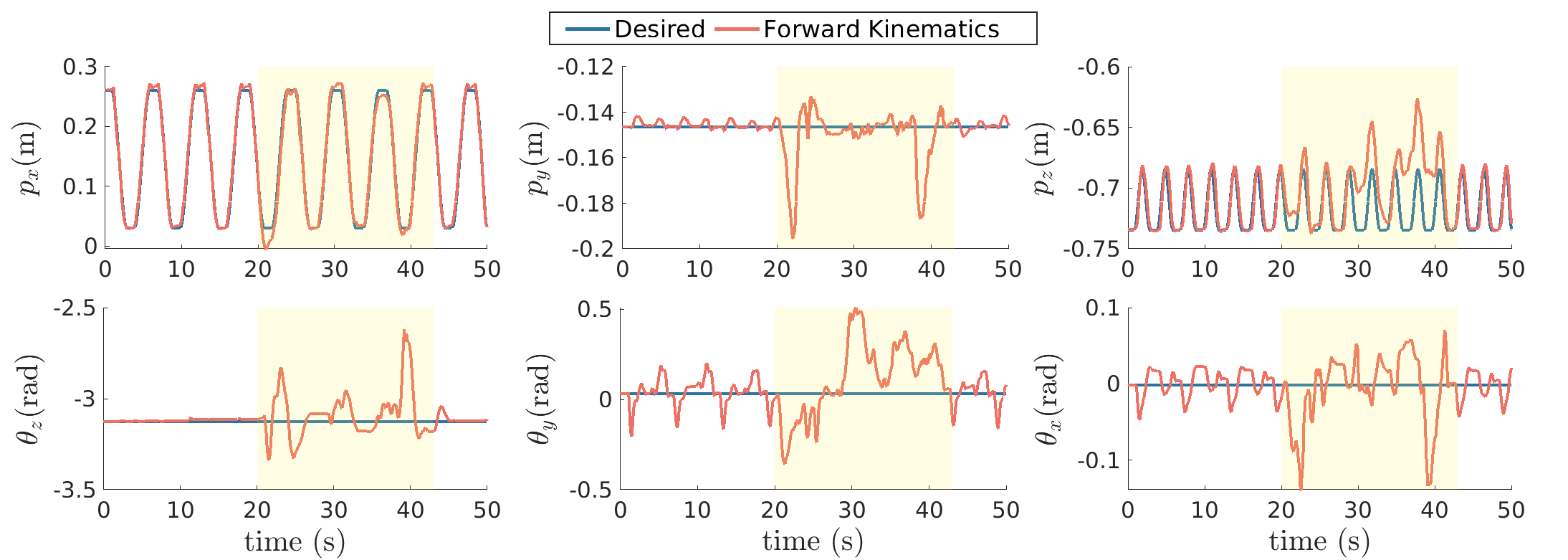}
    \caption{Tracking of a cartesian task defined for the right sole of ergoCub when the leg is subject to external contacts.}
    \label{fig:cartesiantracking}
\end{figure*}

\begin{figure*}[t]
    \centering
    \includegraphics[width=0.95\linewidth]{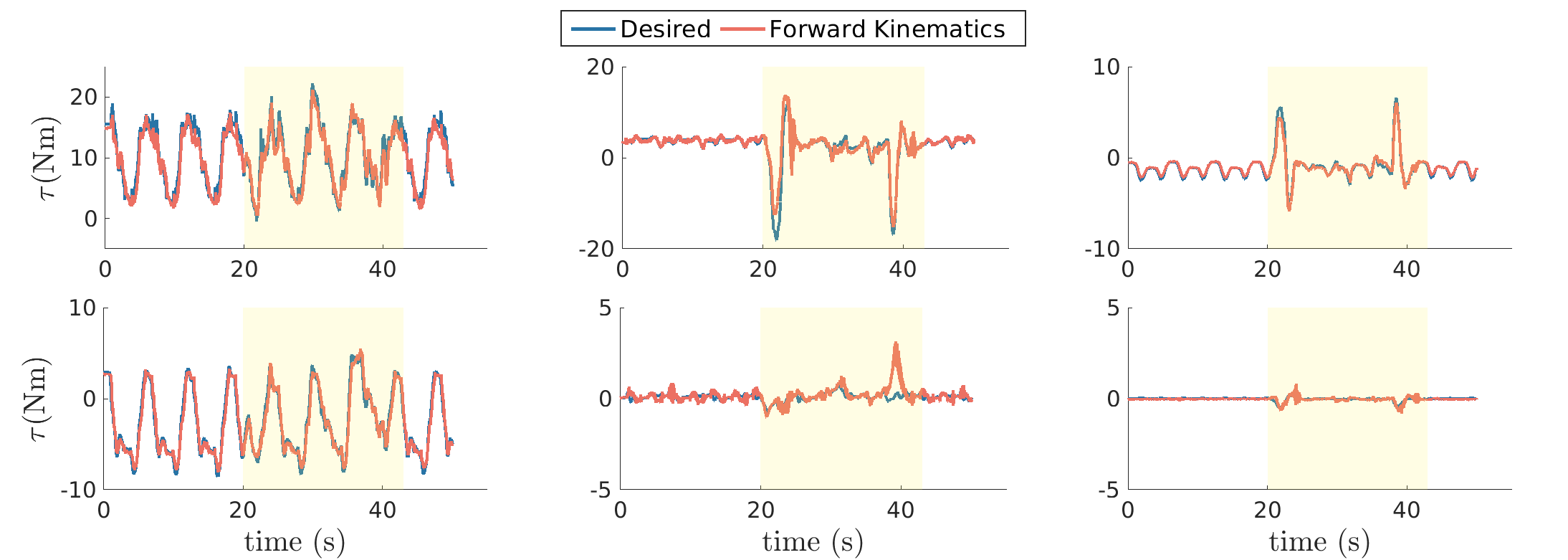}
    \caption{Tracking of the joint torques while performing a cartesian task on the right leg of the ergoCub robot while subject to external contacts.}
    \vspace{-10pt}
    \label{fig:trqtrackingcartesian}
\end{figure*}

\subsubsection{PD control with Gravity Compensation}
For this validation, we conducted two separate experiments. In Experiment 1, the controller has the task of tracking a sinusoidal joint trajectory in one of the robot leg joints. In Experiment 2, we control all three joints of the ergoCub torso with the aim of maintaining their initial configuration. The control method involves PD control with gravity compensation, implemented through the high-level controller described in Section \ref{sec:controller}, by adding a joint position regularization task.

In Experiment 1 the high-level controller generates the necessary torques to achieve the desired joint position, velocity, and acceleration. Figure \ref{fig:pdobstacle} illustrates the torque tracking error for a single joint in the robot leg. In this experiment, an obstacle is placed in front of the leg, preventing it from fully executing the trajectory. The contact point occurs higher than the FT sensor location in the kinematic chain. As in the previous test, the RNEA algorithm cannot measure the external contact, resulting in a large torque tracking error (Root Mean Square Error (RMSE) $=18.3$ Nm). In contrast, our UKF-based approach, integrating current measurements that vary due to external contact, achieves significantly smaller tracking errors (RMSE $ = 1.96$ Nm) than the state-of-the-art solution.

In Experiment 2 we employ the same high-level controller to maintain all three joints of the ergoCub torso in a constant configuration. Random external disturbances are applied, pushing the robot from various contact points. The objective is to prove that the state-of-the-art controller offers a solution that stiffens the robot when contacts occur away from FT sensors. In contrast, our solution ensures compliance with external forces and guarantees the tracking of the desired task once the contact ceases. It is evident from Figure \ref{fig:torso} that in the case of RNEA, the controller becomes unstable due to significant torque tracking errors at the joints.
Conversely, with the UKF-based approach, torques are accurately tracked. Indeed, the RMSE for the three joints are $2.03$ Nm, $0.8$ Nm, and $0.6$ Nm. The multimedia material attached to this paper shows the torso behavior with both algorithms.

\subsection{Tracking of a cartesian task}

In this last validation test, the controller aims to track a specific link pose. The cartesian trajectory is run on a single leg of the robot ergoCub. External forces are applied to perturb the leg to assess the stability of our torque estimation and control. %
Figure \ref{fig:cartesiantracking} shows the desired position and orientation tracking of the controlled frame. The highlighted yellow intervals correspond to time-slots when external disturbances are exerted. We can observe a delay in orientation tracking which can be attributed to static friction in the harmonic drive units which is not modeled~\cite{johanastrom2008revisiting}. As the torque tracking error grows, the torque controller progressively requests higher motor currents overcoming the static friction effect. %
Figure \ref{fig:trqtrackingcartesian} illustrates the desired torque tracking. The RMSE of the torque tracking for the leg joints are $1.31$ Nm, $1.41$ Nm, $0.31$ Nm, $0.64$ Nm, $0.38$ Nm, and $0.08$ Nm.

\section{Conclusions}
\label{sec:conclusions}

This paper contributes to the design of a joint torque estimator for humanoid robots equipped with harmonic drives and without joint torque sensors. We develop a UKF-based sensor fusion algorithm, robust against external disturbances, sensor measurement noises, and modeling errors. We validate our estimation approach on the position-controlled humanoid robot ergoCub, by comparing the proposed algorithm with the state-of-the-art deterministic approach based on RNEA. The validation results show that the novel method ensures effective tracking of the desired torques and high-level tasks, even in the presence of external disturbances.

As a future development, we aim to enhance friction model identification by incorporating more advanced models~\cite{johanastrom2008revisiting}, as well as integrating tactile sensors to improve the estimation of external contact points~\cite{fumagalli2012force}. We also plan to compare our approach with other methods that utilize motor currents and joint and motor encoder measurements for the estimation~\cite{nagamatsu2017distributed}.
Furthermore, we plan to evaluate the performances of our approach in handling more complex whole-body tasks, such as balancing on the ground or walking.

\bibliography{IEEEexample}
\bibliographystyle{IEEEtran}

\end{document}